\documentclass[letterpaper,10pt,conference]{ieeeconf}

\IEEEoverridecommandlockouts
\usepackage{fontspec}
\usepackage{amsmath,amssymb,bm}
\DeclareFontFamily{U}{msa}{}
\DeclareFontFamily{U}{msb}{}
\DeclareFontFamily{OMS}{cmbsy}{}
\DeclareFontShape{OT1}{cmr}{bx}{n}{<0-7.99>cmbx8<8-25>cmbx10}{}
\DeclareFontShape{U}{msa}{m}{n}{<0-25>msam10}{}
\DeclareFontShape{U}{msb}{m}{n}{<0-25>msbm10}{}
\DeclareFontShape{OML}{cmm}{b}{it}{<0-25>cmmib10}{}
\DeclareFontShape{OMS}{cmbsy}{m}{n}{<0-5.99>cmbsy5<6-6.99>cmbsy6<7-7.99>cmbsy7<8-8.99>cmbsy8<9-25>cmbsy10}{}
\DeclareFontShape{OMS}{cmbsy}{b}{n}{<0-5.99>cmbsy5<6-6.99>cmbsy6<7-7.99>cmbsy7<8-8.99>cmbsy8<9-25>cmbsy10}{}
\DeclareFontShape{OMS}{cmsy}{b}{n}{<0-5.99>cmbsy5<6-6.99>cmbsy6<7-7.99>cmbsy7<8-8.99>cmbsy8<9-25>cmbsy10}{}
\DeclareFontShape{OMX}{cmex}{m}{n}{<0-25>cmex10}{}
\DeclareFontShape{OMS}{cmsy}{m}{n}{<0-5.99>cmsy6<6-6.99>cmsy6<7-7.99>cmsy7<8-8.99>cmsy8<9-9.99>cmsy9<10-25>cmsy10}{}
\usepackage{booktabs,graphicx,multirow,balance,xcolor,url}
\ifdefined\SkelWAMShowChanges
  \DeclareRobustCommand{\reviewblue}[1]{{\color{blue}#1}}
  \DeclareRobustCommand{\reviewpurple}[1]{{\color[rgb]{0.5,0,0.5}#1}}
\else
  \DeclareRobustCommand{\reviewblue}[1]{{\color{black}#1}}
  \DeclareRobustCommand{\reviewpurple}[1]{{\color{black}#1}}
\fi
\makeatletter
\@ifpackageloaded{natbib}{}{\let\NAT@parse\undefined}
\makeatother

\ifdefined\SkelWAMSubmission
  \usepackage[draft,bookmarks=false]{hyperref}
\else
  \usepackage[hidelinks]{hyperref}
\fi
\hypersetup{pdfauthor={},pdfsubject={},pdfkeywords={}}

\makeatletter
\AtBeginDocument{%
  \typeout{ICRA2027-PROFILE: paper=\the\paperwidth,\the\paperheight;
  text=\the\textwidth,\the\textheight; columnsep=\the\columnsep;
  bodyfont=\f@size; baseline=\the\baselineskip}%
}
\makeatother

\IEEEoverridecommandlockouts

\IfFileExists{tables/data/en/results_macros.tex}{
  \newcommand{\SkelWAMResultSentence}{SkelWAM achieves 43.3\% success over 1,000 episodes (95\% Wilson CI: 40.3--46.4\%).}

}{
  \providecommand{\SkelWAMResultSentence}{Formal Cross10 evaluation is in progress.}

}

\newcommand{\ours}{\textnormal{SkelWAM}}

\title{\LARGE\bfseries SkelWAM: A Skeleton-Guided World-Action Model for\\
Zero-Shot Cross-Embodiment Manipulation}

\author{%
  Pengjun Niu$^{1,*}$,
  Yujia Xie$^{1,*}$,
  Rui Peng$^{2}$,
  Hang Zhao$^{3,\dagger}$,
  and Ke Liu$^{1,4,\dagger}$\\[0.6em]
  {\small $^{1}$School of Advanced Manufacturing and Robotics,
  Peking University, Beijing, 100871, China}\\
  {\small $^{2}$Feagine}\\
  {\small $^{3}$IIIS, Tsinghua University}\\
  {\small $^{4}$Research Center for Robotics,
  Peking University, Beijing, 100871, China}\\[0.4em]
  {\small $^{*}$Equal contribution
  \qquad $^{\dagger}$Corresponding authors}\\[0.2em]
  {\small
  \texttt{hangzhao@tsinghua.edu.cn}
  \qquad
  \texttt{liuke@pku.edu.cn}}%
}

\begin{document}
\bstctlcite{skelwam:bibstyle}
\IEEEaftertitletext{\begin{minipage}{\textwidth}
  \makeatletter\def\@captype{figure}\makeatother
  \centering
  \includegraphics[width=\textwidth]{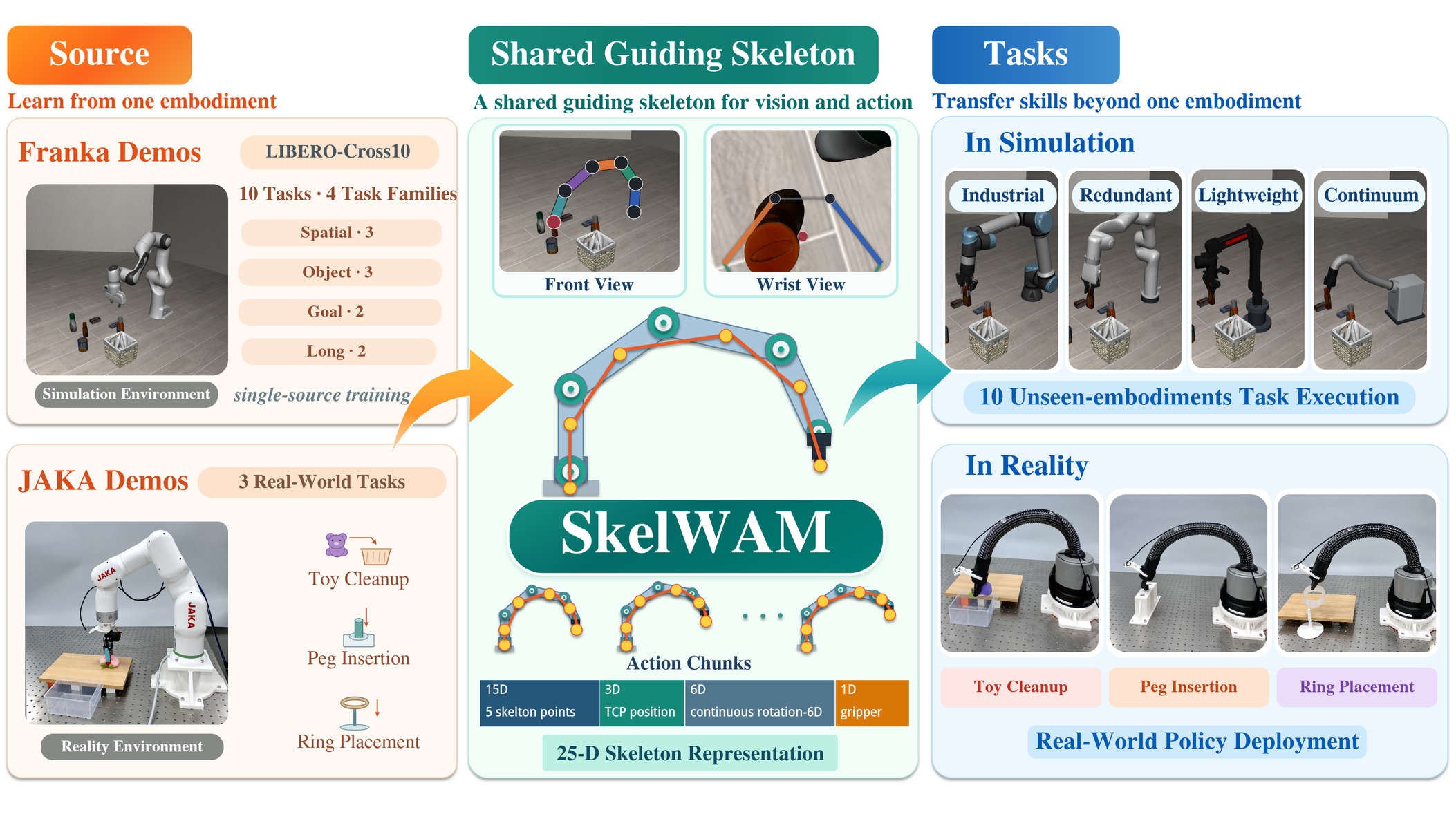}
  \caption{\textbf{SkelWAM overview.}
  The same 25-D centerline/TCP/jaw geometry defines canonical visual
  observations and future action chunks. Embodiment-specific decoders
  translate the predicted whole-body motion intent into native controls
  on LIBERO-Cross10 targets. \reviewblue{Franka demonstrations train the simulation policy.
  In physical experiments, a model trained on JAKA mini2 data is deployed on Feagine A03.}}
  \label{fig:overview}\label{fig:pipeline}
\end{minipage}
\vspace{2pt}}
\maketitle
\thispagestyle{empty}
\pagestyle{empty}

\begin{abstract}
Reusing manipulation experience across robot embodiments is important for
scaling robot learning and reducing repeated task-specific data collection.
However, changes in embodiment alter visual appearance, action dimensionality
and semantics, and the whole-body configurations that can realize the same
tool pose. We present \emph{SkelWAM}, a skeleton-guided world--action model that
couples perception and control through one explicit geometric representation
for single-source cross-embodiment manipulation. Arm centerline geometry,
tool-center-point (TCP) pose, and parallel-jaw commands form a shared 25-D
state. The same definition underlies canonical third-person and wrist
observations and future whole-body action targets. Trained with predictive visual
supervision, a video--action mixture of
transformers predicts canonical skeleton action chunks, which
embodiment-specific constrained decoders convert into joint or
continuum-robot controls. This formulation requires no one-to-one joint
correspondence and uses no target-task demonstrations or target policy updates.
We introduce LIBERO-Cross10, a source-only cross-embodiment transfer benchmark covering 
ten tasks and ten target embodiments across four morphological groups. 
On this benchmark, Franka-trained SkelWAM achieves 43.3\% success over 1,000 episodes,
exceeding the best-performing evaluated baseline by 36.2 percentage points.
\reviewblue{We further deploy a JAKA mini2-trained policy on the Feagine A03 continuum
robot for three tabletop manipulation tasks, illustrating the approach's potential
for real-world cross-embodiment manipulation. Project page: http://www.liukepku.com/skelwam/index.html}
\end{abstract}

\begin{figure*}[t]
  \vspace*{1pt}
  \centering
  \includegraphics[width=\textwidth]{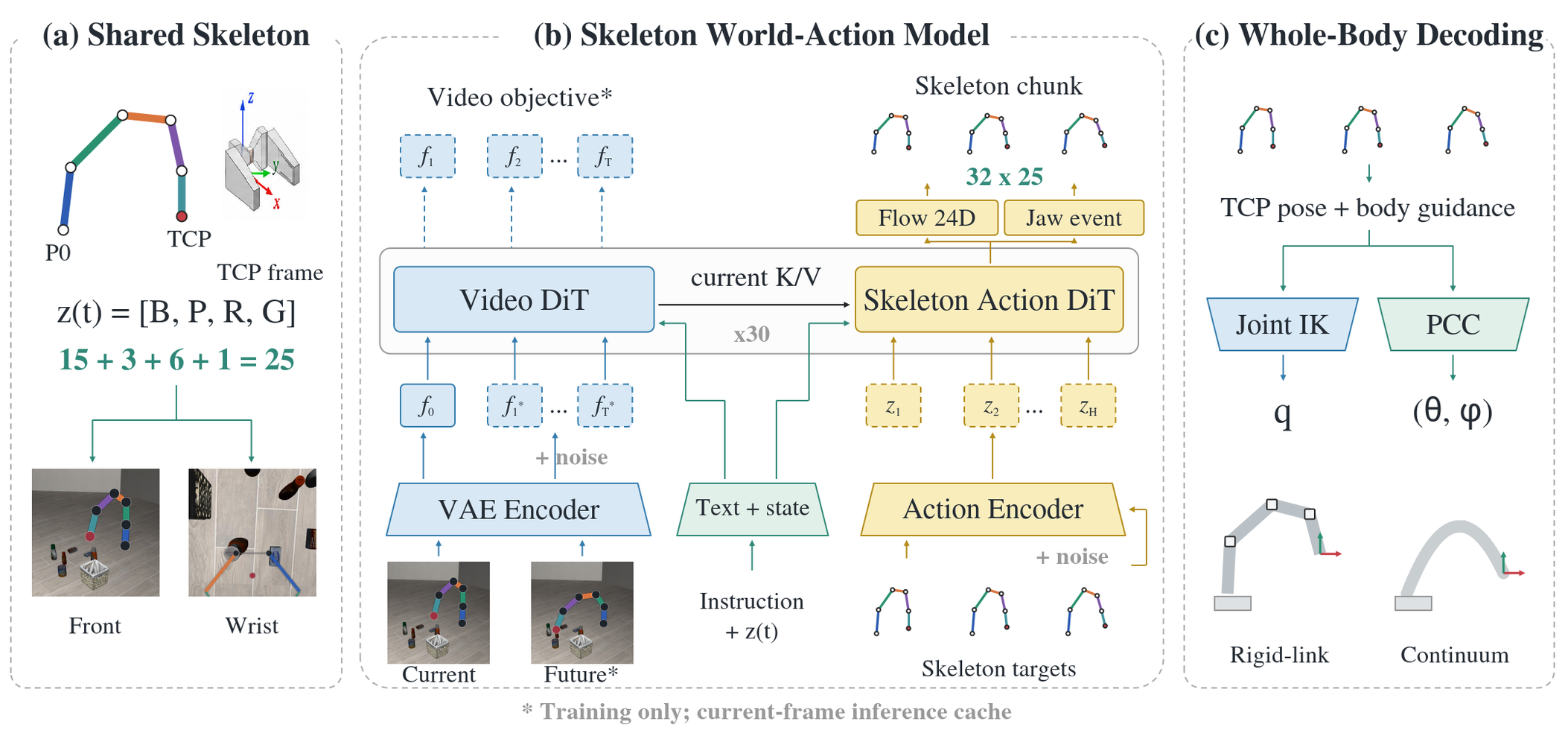}
  \caption{\textbf{SkelWAM world--action model.}
  (a) The shared 25-D body/TCP/jaw state defines external and wrist proxies.
  (b) Thirty paired Video/Action DiT blocks predict a $32\times25$ skeleton
  chunk. \reviewblue{Starred future-video paths provide supervision only during Stage A.
  Action attention accesses current-frame keys and values (K/V), which are cached at
  inference.} The deployment heads combine 24-D continuous flow with a jaw
  event. (c) Native joint IK or PCC decoding prioritizes TCP pose and uses
  the centerline as posture guidance.}
  \label{fig:architecture}
\end{figure*}

\section{INTRODUCTION}
Reusing manipulation experience across robot embodiments can scale robot
learning while reducing repeated task-specific data collection.
World--action models (WAMs) couple predictive visual representations with
action generation, yet deploying a WAM trained on one robot onto another
changes both its observation distribution and its action geometry.
\reviewblue{Multi-robot learning broadens data reuse~\cite{openx2023,ghosh2024octo}.
We study source-only task learning followed by deployment on target
embodiments.} Task demonstrations and policy optimization use one source
robot, with no target-task demonstrations or target policy updates.

This setting introduces two coupled gaps. The visual-side embodiment
gap arises from differences in appearance, body morphology, and
camera-visible geometry. The ction-side embodiment gap includes
different control coordinates, redundancy, and whole-body feasibility across
native kinematic spaces. Visual adaptation can address appearance while
leaving action realization to a controller. End-effector (EEF) or TCP
retargeting supplies a common interaction goal, but the same TCP pose can
correspond to different whole-body configurations. Cross-embodiment WAM
transfer motivates a representation that bridges embodiment
differences on both the visual and action sides.

We introduce \ours{} to connect both sides through a shared skeleton
representation (Fig.~\ref{fig:overview}). Our key idea is to express robot
morphology and motion intent in one canonical whole-body space and use the
same geometry for the visual condition and action target of the WAM. Six
centerline samples, TCP pose, and jaw command define a fixed 25-D canonical
state. This state specifies external and wrist visual proxies and future
skeleton action chunks. \reviewblue{The shared policy predicts whole-body motion intent,
which embodiment-specific decoders realize under each target's kinematic and
execution constraints.}

We introduce LIBERO-Cross10 to evaluate this source-only transfer setting.
The benchmark uses Franka demonstrations from ten selected LIBERO tasks and
ten target embodiments, comprising nine rigid-link arms and one continuum
arm. SkelWAM achieves 43.3\% mean success, compared with 7.1\% for the
best-performing evaluated baseline. Component interventions examine the
shared representation and whole-body execution. \reviewpurple{A qualitative physical study
complements the simulation results with JAKA mini2 source training and
Feagine A03 deployment on three tabletop tasks.}

Our contributions are:
\begin{itemize}
  \item A shared whole-body skeleton representation that canonicalizes visual
  observations and action trajectories with the same geometry across
  embodiments.
  \item A skeleton-guided world--action model that combines predictive visual
  learning with skeleton action generation and embodiment-specific decoding
  for native execution.
  \item LIBERO-Cross10, a ten-task, ten-target source-only transfer benchmark
  spanning diverse rigid-link and continuum embodiments, with 43.3\% mean
  success for SkelWAM.
\end{itemize}

\section{RELATED WORK}
\subsection{Cross-Embodiment Manipulation Policies}
Multi-robot datasets and generalist policies support learning from heterogeneous
platforms~\cite{openx2023,ghosh2024octo}. Visual adaptation methods reduce
appearance shift: Mirage cross-paints a source robot into target observations
~\cite{chen2024mirage}, while RoVi-Aug synthesizes robot and viewpoint
variation during training~\cite{chen2024roviaug}. Shared action representations
include physically interpretable coordinates~\cite{liu2025rdt} and learned
action tokens~\cite{liu2026g05}. UniDex combines hand-masked observations with
functionally aligned actuator coordinates~\cite{zhang2026unidex}.
OPFA learns geometry-aware action latents with a retargeting
decoder~\cite{mu2026opfa}. \reviewblue{Latent Action Diffusion uses contrastive encoders
to align action embeddings and embodiment-specific decoders to recover native actions~\cite{bauer2026lad}.}

\subsection{Structured Representations and World--Action Models}
Motion Tracks uses image-space EEF trajectories~\cite{ren2025motiontracks}.
X-Diffusion uses human-action supervision at high diffusion noise levels
to train policies~\cite{pace2026xdiffusion}. \reviewblue{Robot--object relations support
cross-hand grasping~\cite{wei2025drograsp} and interaction-preserving
retargeting~\cite{wu2026toporetarget}. XMoP uses link-wise geometry for
cross-arm motion planning~\cite{rath2025xmop}.}
RT-Affordance conditions execution on key-stage robot poses~\cite{nasiriany2025rtaffordance},
whereas $A_0$ predicts object-contact points and post-contact trajectories~\cite{xu2025a0}.
$A_1$ accelerates VLA inference with adaptive early exit~\cite{zhang2026a1}.
\reviewblue{World--action models incorporate future-predictive visual
supervision into action learning~\cite{zhu2025uwm,shen2025videovla}. Fast-WAM studies video
co-training with current-frame deployment~\cite{yuan2026fastwam}, and LaWAM
uses compact latent visual subgoals~\cite{chen2026lawam}.} Video models also
synthesize training trajectories~\cite{jang2025dreamgen}.
Soft-arm kinematics can also be learned~\cite{chen2026duallstm}.
SkelWAM focuses on
an explicit centerline/TCP geometry shared by visual conditions and action
targets within a source-trained WAM.

\section{METHOD}
SkelWAM connects the visual and action sides of a world--action model through
shared skeleton geometry. Figure~\ref{fig:architecture} summarizes the
canonical representation, source-only task learning, and embodiment-specific
decoding that maps whole-body motion intent to native controls.

\subsection{Problem Formulation}
We separate source-only task learning from target-specific deployment.
Let $\mathcal D_s=\{(I_t^f,I_t^w,x_t,a_t,\ell)\}$ contain synchronized external
and wrist images, native state and action, and language from one source
embodiment $s$. The target $r$ provides calibrated geometry, cameras, and a
kinematic controller. \reviewblue{Policy optimization and model selection use only source
demonstrations, with target embodiments excluded from the task-training set.}
Encoders $E_r$ and decoders $D_r$ connect native coordinates to a shared policy
$\pi_\theta$ in canonical skeleton space $\mathcal Z$.
Policy weights and source-fitted normalization remain frozen on targets.

\subsection{Canonical Whole-Body Skeleton}
We remove joint-space dependence from the shared policy interface by
representing each embodiment with the same body/TCP/jaw state.
Let $\gamma_r(x_r,u)\in\mathbb R^3$ be the ordered centerline from the mounted
base to the tool center point (TCP), parameterized by normalized arc length
$u\in[0,1]$. We sample $p_i=\gamma_r(x_r,i/5)$ for $i=0,\ldots,5$, with
$p_5$ at the TCP, and define $L_r=\sum_{i=0}^{4}\|p_{i+1}-p_i\|_2$.
Given task-frame normalization $\nu_{\mathcal F}$, the continuous 6-D rotation
representation $\rho(R)$~\cite{zhou2019rotation}, and normalized jaw command
$g_t$, the canonical state is
\begin{equation}
z_t=\left[
\frac{p_0-p_5}{L_r},\ldots,\frac{p_4-p_5}{L_r},
\nu_{\mathcal F}(p_5),\rho({}^{\mathcal F}R_5),g_t
\right]\in\mathbb R^{25}.
\label{eq:skeleton}
\end{equation}
The fields $[B,P,R,G]$ contain 15 body coordinates, 3 TCP position coordinates,
a 6-D orientation, and a jaw command. The six samples follow centerline arc
length rather than joint locations. Expressing body shape
relative to the TCP and normalizing its length separates posture from global
position and scale. The task-frame TCP retains the spatial interaction goal.
Jaw commands $+1/-1$ denote opening/closing. Source-fitted per-coordinate
normalization is applied before policy learning and held fixed at deployment.
This statistical normalization follows the task-coordinate mapping $\nu_{\mathcal F}$.

\subsection{Shared Visual--Action Representation}
\begin{figure*}[t]
  \vspace*{1pt}
  \centering
  \setlength{\parskip}{0pt}
  \setlength{\abovecaptionskip}{-3pt}
  \input{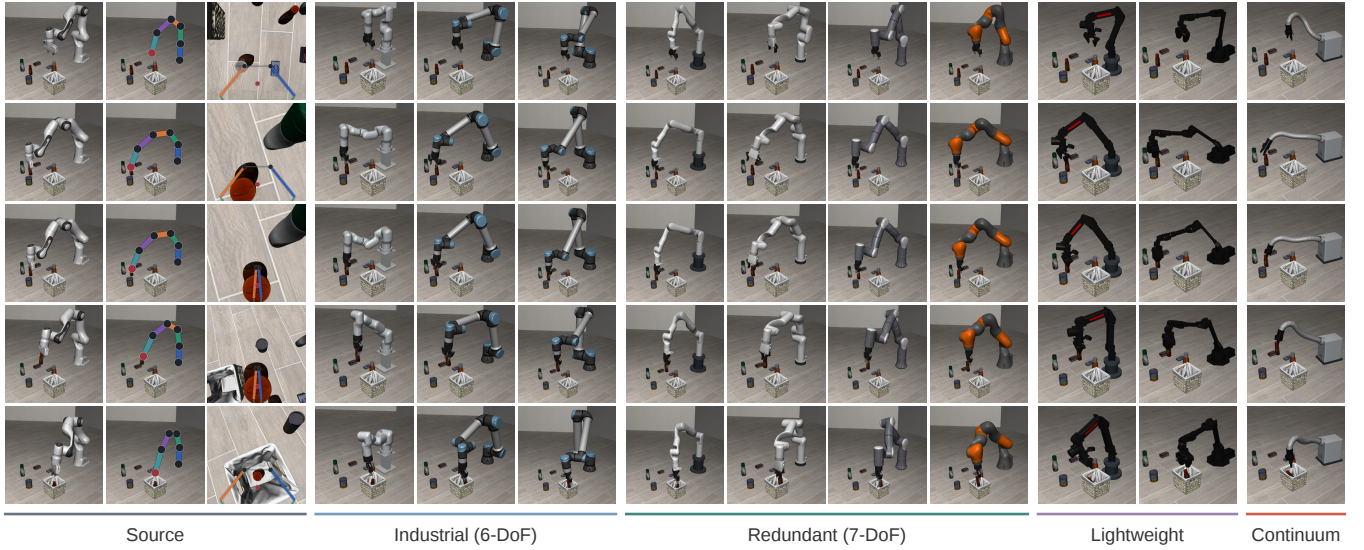}
  \par\nointerlineskip
  \caption{BBQ-sauce-to-basket on Cross-Object (task 3).
  \reviewblue{The first three columns show one Franka source demonstration and its front/wrist
  skeleton views. Target columns follow the robot order in Table~\ref{tab:main}
  and show SkelWAM control replays from recorded states.}
  Rows align initialization, approach, grasp, transport, and the LIBERO goal
  state across independently selected trials.
  \ifnum\figurethreepending=1\relax
  \textit{Working figure: blank columns have not passed complete replay verification.}
  \fi}
  \label{fig:gallery}
\end{figure*}

We use the same skeleton geometry to canonicalize what the policy sees.
We construct canonical visual observations by masking the native robot, filling the
background, and projecting the shared geometry into the image. \reviewblue{The external
view depicts the six-point centerline, while the wrist view depicts a perspective
parallel-jaw skeleton in a fixed canonical TCP-relative camera frame.}
Let $V_r(\cdot;\mathcal C_r)$ denote this visual operator, where $\mathcal C_r$
contains the embodiment-specific calibration parameters. The policy observation is
\begin{equation}
o_t^c=\{V_r(I_t^f,I_t^w,\bar z_t;\mathcal C_r),\bar z_t,\ell\},
\end{equation}
where $\bar z_t=z_t$ for source demonstrations and follows the target feedback
rule below during deployment. The same state definition supplies numerical
policy input, both visual proxies, and future action supervision. One geometry
therefore establishes correspondence between the posture visible to the
policy and the motion coordinates predicted by the policy.
In simulation, segmentation and robot-hidden renders provide the masks and
backgrounds. The physical implementation uses calibrated masks and image
filling (Sec.~\ref{sec:real-robot}). Source images and future skeleton states
are paired along the same demonstration timeline.

The external proxy retains scene context while exposing the arm arrangement
within the workspace. The wrist proxy focuses on local object geometry and
jaw state near the TCP. Their shared construction preserves correspondence
between global posture, local interaction and the predicted action coordinates.

\subsection{Skeleton-Guided World--Action Model}
With both modalities expressed in canonical skeleton space, a WAM learns
their correspondence. \reviewblue{We adopt
paired video--action experts with cached current-frame conditioning and train
the action interface in canonical skeleton space.} \reviewblue{The model pairs a
30-layer Video Expert with a 30-layer Action Expert, both using
diffusion-transformer blocks~\cite{peebles2023dit}.}
\reviewblue{Transformer weights are initialized from the Wan2.2 video backbone in the
Wan model family~\cite{wan2025wan,wan2025wan22}.
The 25-D state encoder, action input/output projections, and gripper-event
heads are newly initialized and trained on source demonstrations.} In each
paired block, action queries attend to current-frame video keys and values
(K/V) and other action tokens. Future-video tokens are excluded from action attention.
The future-video branch supplies an auxiliary predictive objective during
training, while the action attention mask preserves current-observation
conditioning for deployment.

Training has two source-only stages. Stage A jointly adapts the trainable
Video Expert blocks and Action Expert through video prediction and skeleton
action learning:
\begin{equation}
\mathcal L_A=\lambda_v\mathcal L_{\mathrm{video}}+
\mathcal L_{\mathrm{action}}^{25D}.
\end{equation}
The video objective supervises future scene evolution, and the action
objective supervises the associated geometric trajectory.
Stage B freezes the Video Expert and optimizes action generation conditioned
on current-frame features. \reviewblue{Flow matching~\cite{lipman2023flowmatching} predicts 24 geometric coordinates, while a
factorized event head predicts whether the jaw state switches and the time of its first
switch:}
\begin{equation}
\mathcal L_B=\mathcal L_{\mathrm{flow}}^{24D}+
\lambda_g(\mathcal L_{\mathrm{presence}}+\mathcal L_{\mathrm{timing}}).
\end{equation}
Here, $\mathcal L_{\mathrm{presence}}$ and $\mathcal L_{\mathrm{timing}}$ supervise
the occurrence and timing of the first jaw-state change. Separating jaw events
from continuous geometry preserves the discrete opening/closing semantics.
The predictions form a $32\times25$ skeleton action chunk of future absolute
canonical states $\hat Z_{t+1:t+32}$, with at most one jaw-state switch per
chunk. Subsequent switches are handled by replanning.
At inference, the Video Expert encodes the current observation once. Its
cached current-frame K/V condition the Action Expert throughout sampling.
Future-video sampling is unnecessary at deployment.

\subsection{Embodiment-Specific Decoding and Feedback}
\begin{table*}[t]
  \vspace*{5pt}
  \centering
  \caption{Zero-shot success on the ten target embodiments of LIBERO-Cross10 (\%).
  \reviewblue{Columns follow the four target reporting groups, and rows are grouped by method category.}
  Each target has 100 episodes (ten tasks, ten trials per task).}
  \label{tab:main}
  \begingroup
  \setlength{\tabcolsep}{3.5pt}
  \renewcommand{\arraystretch}{1.18}
  \fontsize{8.5}{10}\selectfont
  \resizebox{\textwidth}{!}{
\begin{tabular}{@{}cc*{10}{c}@{\hspace{8pt}}c@{}}
\toprule
\multirow{2}{*}{Family} & \multirow{2}{*}{Method} & \multicolumn{3}{c}{Industrial (6-DoF)} & \multicolumn{4}{c}{Redundant (7-DoF)} & \multicolumn{2}{c}{Lightweight} & \multicolumn{1}{c}{Continuum} & \multirow{2}{*}{Mean} \\
\cmidrule(lr){3-5}\cmidrule(lr){6-9}\cmidrule(lr){10-11}\cmidrule(lr){12-12}
 & & JAKA mini2 & UR5e & UR10e & Kinova Gen3 & xArm7 & Rizon4 & IIWA & ARX-L5 & ViperX300 & Feagine A03 & \\
\midrule
IL & \hyperlink{cite.chi2023diffusion}{Diffusion Policy}~\cite{chi2023diffusion} & 0.0 & 2.0 & 3.0 & 0.0 & 3.0 & 0.0 & 5.0 & 0.0 & 0.0 & \underline{0.0} & 1.3 \\
\addlinespace[3pt]
\multirow{2}{*}{VLA} & \hyperlink{cite.black2025pi05}{$\pi_{0.5}$}~\cite{black2025pi05} & 0.0 & 0.0 & 0.0 & 0.0 & 0.0 & 1.0 & 0.0 & 0.0 & 0.0 & \underline{0.0} & 0.1 \\
 & \hyperlink{cite.kim2025openvlaoft}{OpenVLA-OFT}~\cite{kim2025openvlaoft} & 0.0 & 0.0 & 0.0 & 0.0 & 1.0 & 0.0 & 0.0 & 1.0 & 0.0 & \underline{0.0} & 0.2 \\
\addlinespace[3pt]
\multirow{2}{*}{WAM} & \hyperlink{cite.yuan2026fastwam}{FastWAM}~\cite{yuan2026fastwam} & 0.0 & 0.0 & 0.0 & 0.0 & 4.0 & 0.0 & 0.0 & 0.0 & 0.0 & \underline{0.0} & 0.4 \\
 & \hyperlink{cite.chen2026lawam}{LaWAM}~\cite{chen2026lawam} & 0.0 & 0.0 & 1.0 & 4.0 & 2.0 & 0.0 & 0.0 & 0.0 & \underline{2.0} & \underline{0.0} & 0.9 \\
\addlinespace[3pt]
\multirow{2}{*}{X-Emb.} & \hyperlink{cite.chen2024mirage}{Mirage}~\cite{chen2024mirage} & 0.0 & 0.0 & 0.0 & 0.0 & 4.0 & 0.0 & 0.0 & 0.0 & 0.0 & \underline{0.0} & 0.4 \\
 & \hyperlink{cite.chen2024roviaug}{RoVi-Aug}~\cite{chen2024roviaug} & \underline{3.0} & \textbf{15.0} & \underline{4.0} & \underline{7.0} & \underline{13.0} & \underline{8.0} & \underline{9.0} & \underline{10.0} & \underline{2.0} & \underline{0.0} & \underline{7.1} \\
\midrule
Ours & \textbf{SkelWAM} & \textbf{59.0} & \underline{13.0} & \textbf{62.0} & \textbf{49.0} & \textbf{19.0} & \textbf{62.0} & \textbf{22.0} & \textbf{56.0} & \textbf{32.0} & \textbf{59.0} & \textbf{43.3} \\
\bottomrule
\end{tabular}
}
  \par\vspace{3pt}
  \begin{minipage}{\textwidth}\footnotesize
  \reviewblue{\textbf{Bold} and \underline{underlined} values indicate the highest and second-highest distinct rates
  in each column, respectively, including ties.} Mean is the
  equally weighted average over ten targets. \reviewblue{Non-SkelWAM methods use their
  native policy interface with target-specific EEF-IK. Mirage and RoVi-Aug use
  FastWAM as the underlying policy.} Ablations are reported separately.
  \end{minipage}
  \endgroup
\end{table*}

\begin{figure*}[t]
  \vspace*{1pt}
  \centering
  \includegraphics[width=\textwidth,trim=0 28bp 0 0,clip]{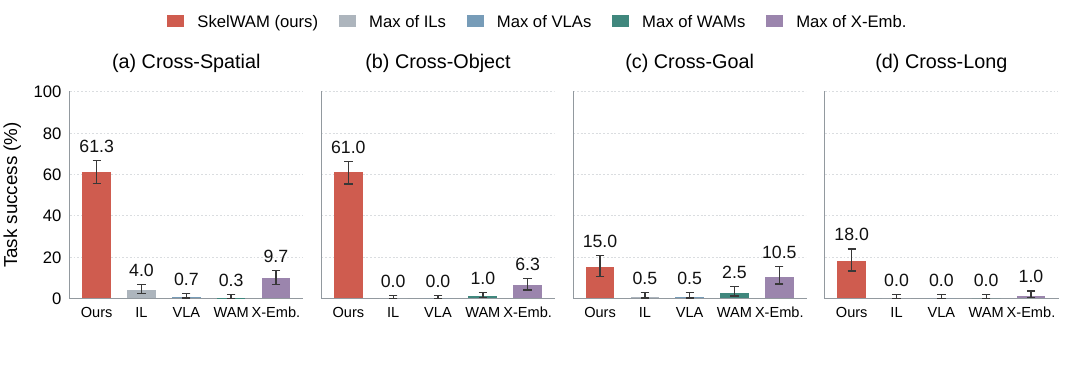}
  \caption{Success on the four selected task subsets, averaged over ten target
  embodiments. Each baseline bar is the maximum complete-method average within
  its family. 
  Error bars show descriptive, episode-level Wilson 95\% confidence intervals
  from binary outcomes (300, 300, 200, and 200 episodes per displayed method in
  the four panels), without adjustment for maximum selection. Task subsets
  are defined in the benchmark section.}
  \label{fig:results}
\end{figure*}

\reviewblue{The policy is shared across embodiments, while native execution remains
embodiment-specific.}
For desired state $\hat z$, decoder $D_r$ prioritizes TCP pose and jaw commands,
subject to native coordinate limits and feasibility checks. In addition to its
EEF objective, it uses the posture and continuity regularizer
\begin{equation}
\sum_{i=0}^{4}w_i\|f_{r,i}(x_r)-\hat p_i\|_2^2+
\lambda_q\|x_r-x_{r,t-1}\|_2^2,
\label{eq:decoder}
\end{equation}
where $f_{r,i}(x_r)$ are target-kinematics centerline samples and $\hat p_i$
are reference points decoded at the target scale. Simulation decoding rescales
body offsets with the target's sampled centerline length at decoder initialization
and adds the task-frame TCP position. \reviewblue{The rigid-link decoder uses constrained
joint IK, whereas the continuum decoder operates on piecewise-constant-curvature (PCC)
parameters~\cite{webster2010continuum} and available tool coordinates.} The
TCP term anchors task interaction, and the centerline term provides soft
whole-body posture guidance subject to target feasibility. Exact centerline
agreement is not required. Continuity discourages abrupt changes between
successive native commands. Redundant arms can redistribute joint motion,
while the continuum arm realizes the reference through segment curvature.

The policy state combines canonical intent with interaction feedback.
The simulation bridge uses measured geometry when no prior intent exists.
For rigid-link targets, the previous body intent is retained while measured
TCP position, orientation, and jaw command update the next policy state. For
the continuum target, body and interaction-orientation intent are retained,
with measured TCP position and jaw feedback. The decoder uses the measured
native configuration for its kinematic and feasibility computations. This
separates the policy's posture reference from the target's measured body
shape. Execution is receding-horizon:
after the configured prefix is applied, fresh observations update the state
and initiate a new chunk.

\section{EXPERIMENTS}
We evaluate source-only cross-embodiment transfer on LIBERO-Cross10,
compare against representative policies, and examine the roles of the shared
representation and whole-body decoding. \reviewpurple{We also deploy a JAKA mini2-trained
policy on a physical Feagine A03 continuum arm.}

\subsection{LIBERO-Cross10 Benchmark}
We introduce LIBERO-Cross10 to evaluate source-trained manipulation policies
under changes in robot morphology. Built on LIBERO~\cite{liu2023libero}, it
retains objects, language instructions, object reset states, and success
predicates while registering target-specific arms, grippers, installations,
and cameras. \reviewblue{We use a fixed set of ten tasks: task IDs 0, 2, and 3 from both Spatial and Object,
and task IDs 5 and 8 from both Goal and LIBERO-10.}
We refer to these selected subsets as Cross-Spatial, Cross-Object, Cross-Goal,
and Cross-Long, respectively.
Franka Panda is the source and is excluded from all ten-target averages.
\reviewblue{We report results for four fixed, non-overlapping groups:
\emph{Industrial} (6-DoF: JAKA mini2, UR5e, and UR10e),
\emph{Redundant} (7-DoF: Kinova Gen3, xArm7, Rizon4, and IIWA),
\emph{Lightweight} (ARX-L5 and ViperX300),
and \emph{Continuum} (Feagine A03).}
The first three groups contain nine rigid-link arms. Feagine A03 supplies a continuum target with
PCC kinematics. All targets use parallel-jaw grippers with their native or
shared contact geometries. The benchmark uses MuJoCo~\cite{todorov2012mujoco}
and robosuite~\cite{zhu2020robosuite}. Figure~\ref{fig:gallery} illustrates
the common task across the source and target roster.

\subsection{Evaluation Protocol}
\begin{table}[t]
  \vspace*{5pt}
  \centering
  \caption{Component interventions on the same 1,000 episodes per variant.
  $\Delta$ is ablation minus full, in percentage points.}
  \label{tab:ablations}
  \begingroup
  \setlength{\tabcolsep}{3.5pt}\renewcommand{\arraystretch}{1.18}
  \fontsize{8.5}{10}\selectfont
\begin{tabular*}{\linewidth}{@{\extracolsep{\fill}}c*{4}{c}cc@{}}
\toprule
Variant & $V$ & $A$ & $W$ & $D$ & SR (\%) & $\Delta$ (pp) \\
\midrule
\textbf{Full} & $\checkmark$ & $\checkmark$ & $\checkmark$ & $\checkmark$ & \textbf{43.3} & -- \\
w/o $V^{R}$ & $\times$ & $\checkmark$ & $\checkmark$ & $\checkmark$ & 0.0 & -43.3 \\
w/o $A^{R}$ & $\checkmark$ & $\times$ & $\checkmark$ & $\times$ & 0.1 & -43.2 \\
w/o $W^{R}$ & $\checkmark$ & $\checkmark$ & $\times$ & $\checkmark$ & \underline{43.0} & -0.3 \\
w/o $D^{I}$ & $\checkmark$ & $\checkmark$ & $\checkmark$ & $\times$ & 40.9 & -2.4 \\
\bottomrule
\end{tabular*}

  \par\vspace{3pt}
  \begin{minipage}{\columnwidth}\footnotesize
  \reviewblue{$V$, $A$, $W$, and $D$ denote skeleton vision, body-action supervision, wrist input,
  and the decoder body objective, respectively.} \reviewblue{$R$ denotes source retraining, and $I$ denotes an inference-only
  intervention using the full checkpoint.} Removing $A$ also disables $D$,
  coupling changes in learning and execution.
  \end{minipage}\endgroup
\end{table}

\begin{figure}[t]
  \vspace*{1pt}
  \centering
  \includegraphics[width=\columnwidth]{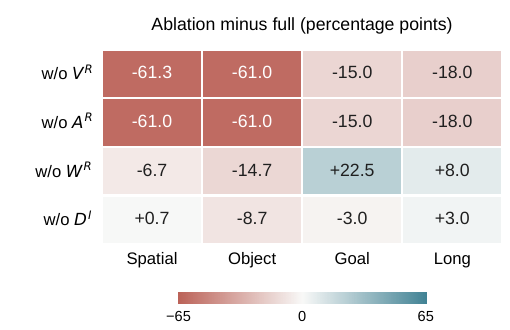}
  \caption{Task-dependent effects of the interventions in Table~\ref{tab:ablations}.
  Cells show ablation minus full in percentage points on the four Cross-* subsets.
  Red indicates a decrease and blue an increase in success.}
  \label{fig:ablation-effects}
\end{figure}

\begin{figure}[!tb]
  \vspace*{1pt}
  \centering
  \setlength{\abovecaptionskip}{0pt}
  \includegraphics[width=\columnwidth]{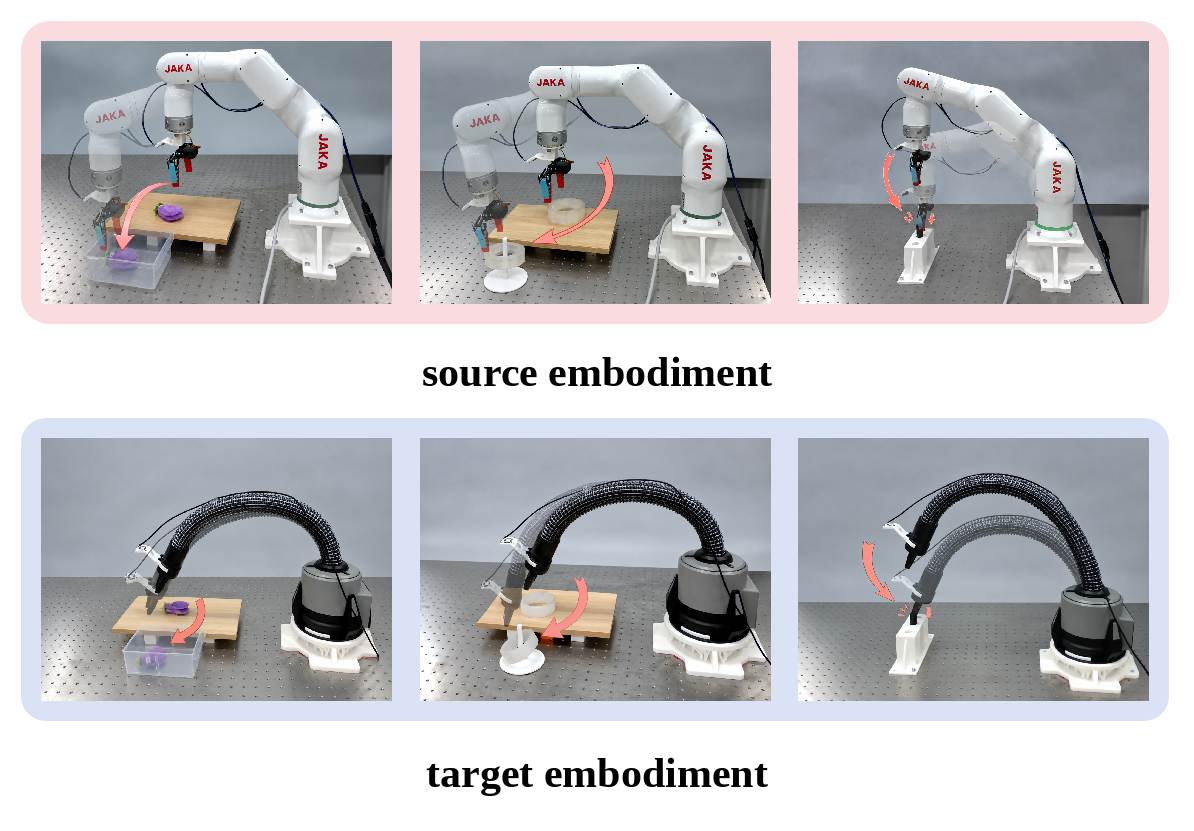}
  \caption{\textbf{Real-robot task setup.} \reviewblue{Columns show Toy Cleanup, Ring
  Placement, and Peg Insertion, while rows show source JAKA mini2 and target
  Feagine A03.} Arrows and translucent poses illustrate the task motions.}
  \label{fig:real-setup}
\end{figure}

\begin{figure*}[!t]
  \vspace*{1pt}
  \centering
  \input{figures/fig07_real_implementation/crop}
  \makebox[0.43\textwidth]{\small\bfseries JAKA mini2: Data Collection}%
  \makebox[0.43\textwidth]{\small\bfseries Feagine A03: Policy Deployment}\par
  \vspace{3pt}
  \includegraphics[width=0.86\textwidth,trim=0 0 0 \figsevenheadercrop,clip]{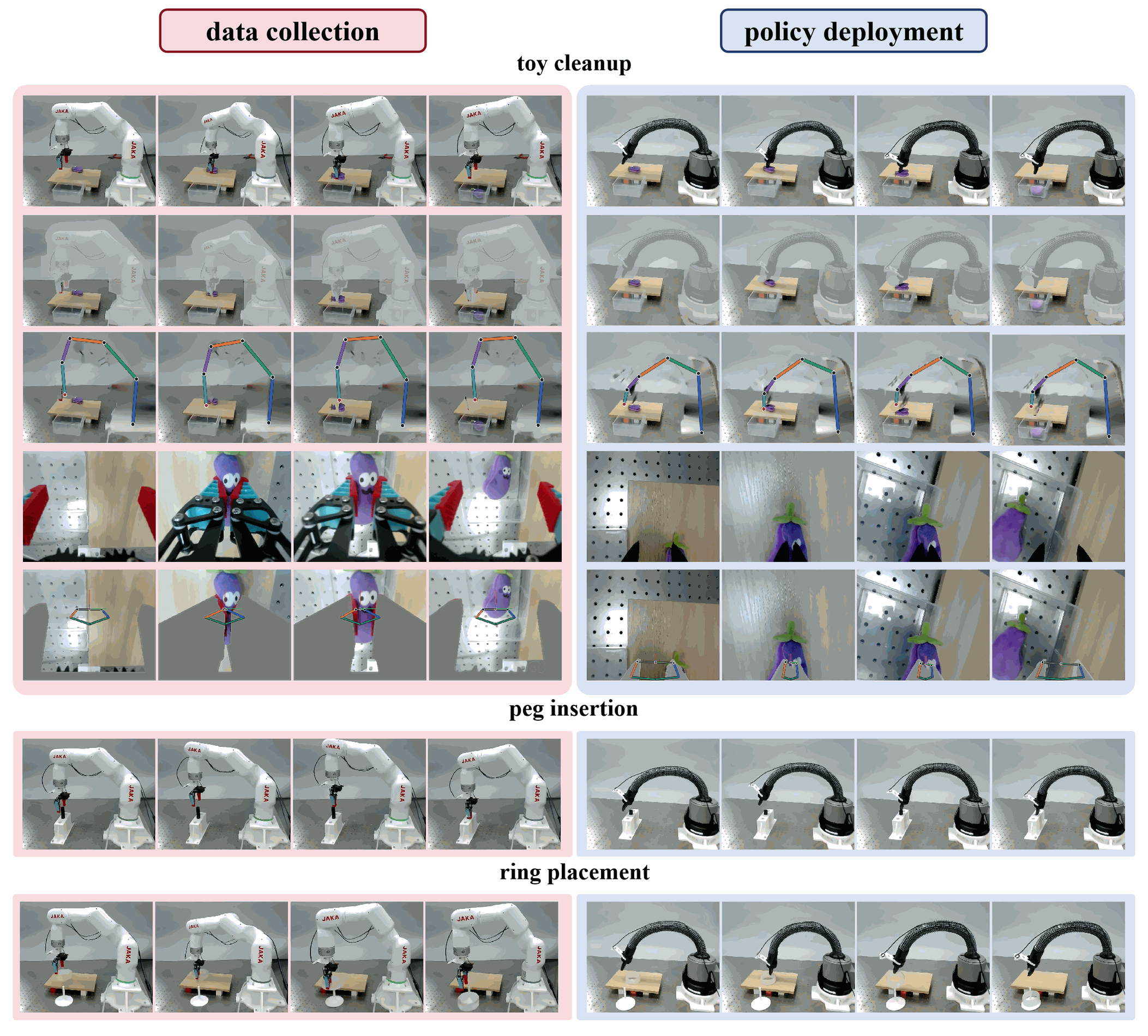}
  \par\nointerlineskip
  \setlength{\abovecaptionskip}{3pt}
  \caption{\textbf{Physical visual--action interface.}
  \reviewblue{JAKA mini2 is shown on the left and Feagine A03 on the right.} For Toy Cleanup, the five rows show
  external RGB, mask overlays, skeleton composites, wrist RGB, and wrist
  illustrations. Lower bands: Peg Insertion and Ring Placement. Each side shows
  four selected times.}
  \label{fig:real}
\end{figure*}

\textbf{Source training.}
The source set contains 467 successful Franka demonstrations from the ten
selected tasks, split by episode into 447 training and 20 validation episodes.
Each window contains two $224\times224$ canonical views, language, the current
25-D state, and 32 future absolute states. Video targets are sampled at
offsets $0,4,\ldots,32$. \reviewblue{Stage A trains the final four video blocks, video
output head, and Action Expert for 6k updates. Stage B then freezes the Video Expert and
trains the action pathway for 24k updates.} Model selection uses held-out
Franka episodes and selects Stage-B step 23k after the 6k Stage-A updates.
Inference uses 20 flow-matching steps and executes ten predicted states before
replanning.

\textbf{Baselines and adaptation.}
\reviewblue{We evaluate four method families: imitation learning (Diffusion Policy~\cite{chi2023diffusion}),
vision--language--action policies ($\pi_{0.5}$~\cite{black2025pi05} and
OpenVLA-OFT~\cite{kim2025openvlaoft}), world--action models
(Fast-WAM~\cite{yuan2026fastwam} and LaWAM~\cite{chen2026lawam}), and
cross-embodiment visual adaptation (Mirage~\cite{chen2024mirage} and
RoVi-Aug~\cite{chen2024roviaug}).} The last two are adaptations of these
approaches to a common Fast-WAM backbone.

\reviewblue{Diffusion Policy is source-trained, whereas the VLA and WAM baselines use released
checkpoints.} Mirage reuses Fast-WAM weights with visual cross-painting,
whereas RoVi-Aug uses source-data visual augmentation and a 4k adapted
checkpoint. Pretraining and source-adaptation budgets differ across these methods.
Each baseline retains its method-specific image processing and native action
contract. \reviewblue{Cartesian outputs pass through target-specific EEF inverse kinematics
in joint or PCC coordinates (EEF-IK). No baseline
receives SkelWAM skeleton images, centerline state, or a body-tracking
objective.} Results compare deployed systems, including their adapters.

\textbf{Trials and aggregation.}
Every method uses the same ten targets, ten tasks, and ten paired trial
indices per embodiment--task cell. Let $\mathcal{R}$ and $\mathcal{T}$ denote
the target-embodiment and task sets, respectively, and let $J$ be the number
of trials per cell. Let $y_{r,k,j}=1$ when trial $j$ on embodiment $r$ and
task $k$ satisfies the original LIBERO success predicate, and $0$ otherwise.
\begin{equation}
\mathrm{SR}=
\frac{1}{|\mathcal{R}|\,|\mathcal{T}|\,J}
\sum_{r\in\mathcal{R}}\sum_{k\in\mathcal{T}}\sum_{j=1}^{J}y_{r,k,j}.
\end{equation}
We report SR as a percentage. Here,
$|\mathcal{R}|=|\mathcal{T}|=J=10$, giving 1,000 episodes per method.
\reviewblue{All embodiment--task cells are equally weighted, with the four task families
contributing 300, 300, 200, and 200 episodes.} The paired trial indices use
the same initializations across methods. Error bars use 95\% Wilson intervals
for pooled episode proportions. Zero-shot evaluation fixes the policy after
source training and permits task-independent target calibration.

\subsection{Zero-Shot Transfer Results}
\textbf{Comparison with prior methods.}
Table~\ref{tab:main} compares zero-shot success across the complete target
matrix. \SkelWAMResultSentence{} \reviewblue{The best-performing evaluated baseline,
the Fast-WAM-based RoVi-Aug implementation with EEF-IK, reaches 7.1\%,
giving SkelWAM an absolute improvement of 36.2 percentage points.} Fast-WAM and its
Mirage cross-painted variant both obtain 0.4\%.

\reviewblue{SkelWAM achieves the highest success rate on nine targets. On UR5e,
RoVi-Aug obtains 15.0\% compared with 13.0\% for SkelWAM.}
Figure~\ref{fig:results} reports the highest average success within each
baseline family for each task subset, with every method averaged over all ten
target embodiments.

\textbf{Task and morphology analysis.}
Figure~\ref{fig:results} shows 61.3\% and 61.0\% success on Cross-Spatial and
Cross-Object, versus 15.0\% and 18.0\% on Cross-Goal and Cross-Long. Transfer
performance varies substantially with the task subset despite the shared
interface.

The target matrix also spans different embodiment groups: the mean success
rates are 44.7\% for Industrial, 38.0\% for Redundant, 44.0\% for Lightweight,
and 59.0\% for the single Continuum target. Variation within groups includes the
19.0\%--62.0\% range among the seven-DoF arms.

\subsection{Component Analysis}
\textbf{Ablation design.}
Table~\ref{tab:ablations} examines four components: skeleton
vision ($V$), body-action supervision ($A$), wrist input ($W$), and the decoder's
body objective ($D$). The $V$, $A$, and $W$ variants are separately trained on
Franka data. The $D$ intervention alone reuses the selected full-model weights.
All scores use the same 100 target-embodiment--task cells and ten paired trial indices.

\textbf{Representation and execution.}
The w/o $V^{R}$ and w/o $A^{R}$ variants obtain 0.0\% and 0.1\%, respectively.
The action variant retains the 25-D architecture, removes supervision on its
15 body coordinates, and uses an EEF-only decoder objective. \reviewblue{This intervention
changes both learning and execution, whereas removing $D$ changes execution alone.}
The visual variant is trained with native source RGB.

\textbf{Task-dependent effects.}
Removing the wrist changes overall success by only $-0.3$ points, yet its
effects are $-6.7/-14.7/+22.5/+8.0$ points across the four task subsets.
Figure~\ref{fig:ablation-effects} reveals these task-dependent reversals.
Removing only the decoder body objective changes the mean by $-2.4$ points. Its effects are
also task dependent: $-8.7$ on Cross-Object but $+3.0$ on Cross-Long.
Thus, the wrist and whole-body objective contribute differently across task
families.

\subsection{Real-Robot Deployment}
\label{sec:real-robot}

\reviewpurple{\textbf{Setup and source training.}}
JAKA mini2 and Feagine A03 each use external and wrist RGB cameras for Toy
Cleanup, Ring Placement, and Peg Insertion (Fig.~\ref{fig:real-setup}).
The 277 quality-controlled JAKA mini2 teleoperation demonstrations are split by episode
into 249 training and 28 validation trajectories, then compiled into 10\,Hz
skeleton windows. \reviewpurple{We train
the Action Expert and final two video blocks on JAKA mini2 data for 6k updates
with the action objective.} Model selection uses source validation only.

\textbf{Physical interface.}
Calibrated JAKA mini2 forward kinematics and Feagine A03 PCC models encode native
feedback into the shared 25-D representation. Task-frame, tool, and camera calibration
registers the numerical skeleton with both camera views. The TCP is defined
at the gripper contact center, accounting for the different tool lengths.
\reviewblue{Embodiment-specific constrained inverse solvers map predicted skeleton trajectories
to native commands. Feagine A03 uses its PCC decoder with interaction feedback.}
For vision, classical computer vision combines static-ROI feature matching
with RANSAC, forward--backward Lucas--Kanade flow, and color-assisted TCP
relocking for masking and skeleton-point tracking.

\textbf{Three-task deployment.}
\reviewpurple{We deploy the JAKA mini2-trained policy on Feagine A03 with fixed weights and a
target-specific PCC decoder for Toy Cleanup, Ring Placement, and Peg
Insertion.} No Feagine A03 task demonstrations are used for fine-tuning.
Figure~\ref{fig:real} shows the qualitative deployments and the associated
external and wrist-view processing.

\reviewpurple{Transferring the JAKA mini2-trained policy to Feagine A03 illustrates
SkelWAM's potential for real-world cross-embodiment manipulation.}

\section{DISCUSSION AND LIMITATIONS}
The results support shared visual--action geometry as a backbone for transferring
source-trained WAMs across embodiments. Cross-Goal and Cross-Long remain lower
than Cross-Spatial and Cross-Object, and the ablations show that wrist
context and posture guidance interact with the manipulation objective.
\reviewblue{The comparison uses a common target protocol, although released pretraining and
source adaptation differ across methods. The coupled action intervention also
limits component-level attribution.}

Transfer depends on reachability, gripper contact geometry, camera calibration,
and the quality of background filling under object occlusion. These factors
remain in the target interface.

\section{CONCLUSION}
Cross-embodiment world--action model (WAM) transfer faces visual-side and action-side embodiment
gaps. SkelWAM connects both through a shared 25-D whole-body skeleton: the
same geometry defines visual conditions, canonical state, and future skeleton
action chunks. A source-trained WAM predicts motion intent, and
embodiment-specific decoders realize it under native constraints.
On LIBERO-Cross10, Franka-trained SkelWAM achieves 43.3\% mean success across
ten targets, including nine rigid-link and one continuum embodiment, 36.2
percentage points above the best-performing evaluated baseline. The
qualitative JAKA mini2-to-Feagine A03 study demonstrates physical deployment
of the shared interface.

\bibliographystyle{./IEEEtran}
\setlength{\itemsep}{0.25ex plus 0.1ex}
\balance
\bibliography{references/bibliography_style,references/references}
\end{document}